\documentclass[10pt,twocolumn,letterpaper]{article}

\usepackage{wacv}              
\usepackage{svg}

\usepackage[normalem]{ulem}
\definecolor{wacvblue}{rgb}{0.21,0.49,0.74}
\usepackage[pagebackref,breaklinks,colorlinks,allcolors=wacvblue]{hyperref}
\usepackage{booktabs}
\usepackage{multirow}
\usepackage{siunitx}
\usepackage{bm} 
\usepackage{tcolorbox} 
\usepackage[table]{xcolor}
\def\wacvPaperID{2679} 
\def\confName{WACV}
\def\confYear{2026}

\title{SmokeBench: Evaluating Multimodal Large Language Models for Wildfire Smoke Detection}

\author{Tianye Qi \qquad
Weihao Li \qquad
Nick Barnes \\[0.5ex]
Australian National University 
}

\begin{document}
\maketitle
\begin{abstract}
Wildfire smoke is transparent, amorphous, and often visually confounded with clouds, making early-stage detection particularly challenging. In this work, we introduce a benchmark, called SmokeBench, to evaluate the ability of multimodal large language models (MLLMs) to recognize and localize wildfire smoke in images. The benchmark consists of four tasks: (1) smoke classification, (2) tile-based smoke localization, (3) grid-based smoke localization, and (4) smoke detection. We evaluate several MLLMs, including Idefics2, Qwen2.5-VL, InternVL3, Unified-IO 2, Grounding DINO, GPT-4o, and Gemini-2.5 Pro. Our results show that while some models can classify the presence of smoke when it covers a large area, all models struggle with accurate localization, especially in the early stages. Further analysis reveals that smoke volume is strongly correlated with model performance, whereas contrast plays a comparatively minor role. These findings highlight critical limitations of current MLLMs for safety-critical wildfire monitoring and underscore the need for methods that improve early-stage smoke localization.
\end{abstract}

\section{Introduction}
In recent years, the frequency and intensity of wildfires have risen dramatically across the globe, posing severe threats to ecosystems, human health, and infrastructure. Recognizing this escalating risk, governments and international bodies, including the United Nations and the World Meteorological Organization, have underscored the necessity of integrating wildfire risk into broader climate adaptation and disaster preparedness strategies \cite{brys2025machine}. Rapid advances in artificial intelligence (AI) have positioned data-driven methods as powerful tools for addressing these challenges \cite{dewangan2022figlib,Li2026AusSmoke,Zhao2026FalseAlarm,Shrestha2025SDIPaste,Shrestha2025GIMO}. Current research spans a diverse set of approaches, ranging from weather–fire interaction modeling to UAV- and IoT-enabled monitoring systems \cite{chan2024iotwildfire}, as well as deep learning techniques for smoke detection in images \cite{yan2022transmission}. These developments demonstrate the potential of AI to enhance early warning systems, enable localized detection, and support adaptive response strategies, ultimately contributing to more resilient wildfire management in the context of climate change.

\begin{figure}[t]
  \centering
  \includegraphics[width=0.9\linewidth]{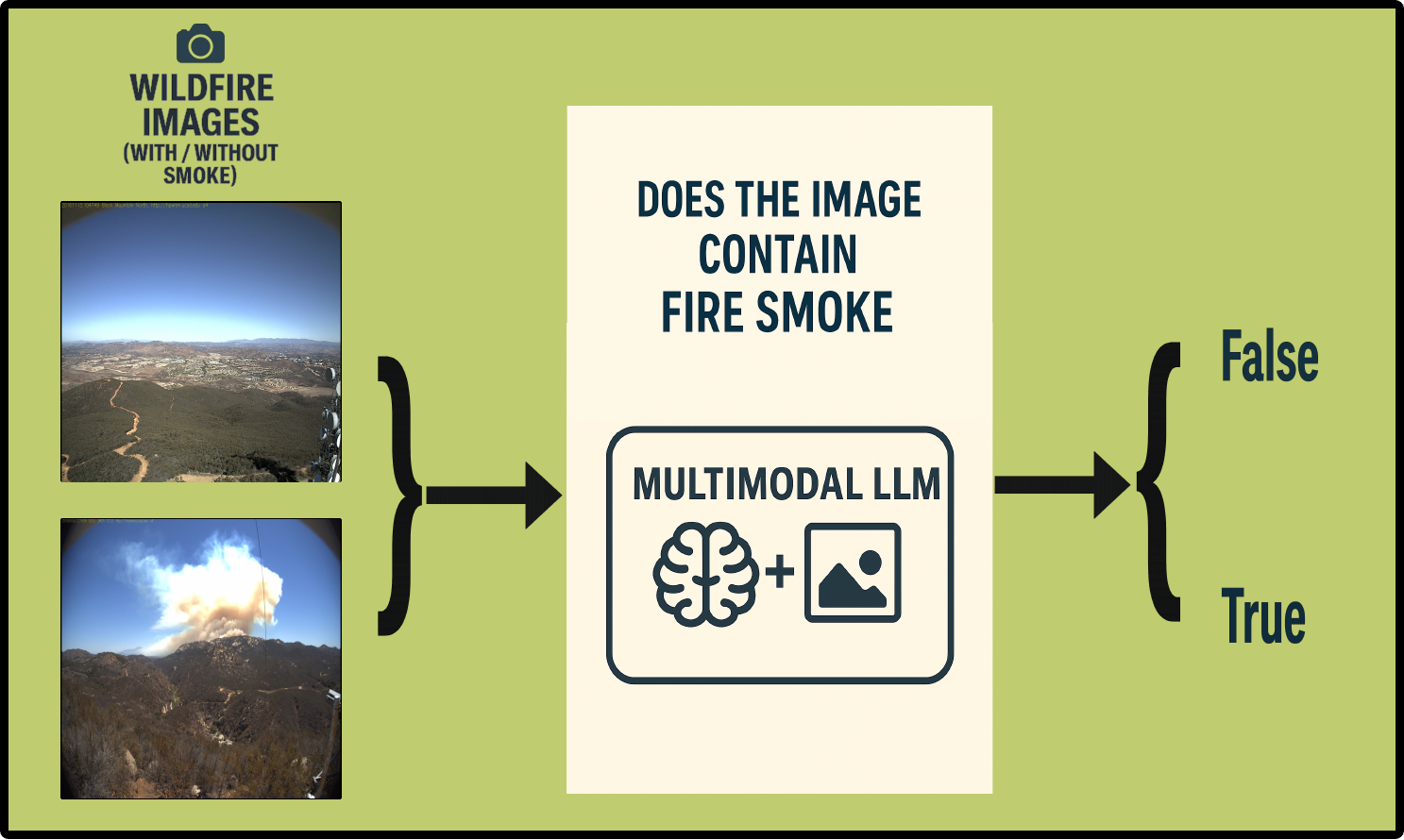}
  \caption{We present SmokeBench, a benchmark to evaluate the ability of multimodal large language models (MLLMs) to detect wildfire smoke in images.
  }
  \label{fig:mllm-smoke}
\end{figure}

Detecting wildfire smoke is particularly challenging due to its transparent and amorphous characteristics, and it has therefore attracted increasing research attention. Several studies have proposed deep learning models specifically for smoke detection~\cite{al2023early, kim2023forest, saydirasulovich2023improved, yan2022transmission}, and multiple dedicated datasets~\cite{dewangan2022figlib, cheng2019smoke, pyro-sdis, lostanlen2024scrapping, yan2022transmission} have been released to support this line of research. Although multimodal large language models (MLLMs) have recently been benchmarked on a wide range of vision–language tasks~\cite{li2024seed, liu2024mmbench, fu2025mme, duan2024vlmevalkit, robicheaux2025roboflow100,Zhao2026DermEVAL}, none of these efforts address wildfire scenarios, which are inherently safety-critical and require reliable early detection. To the best of our knowledge, the potential of MLLMs for wildfire smoke detection has not yet been explored, leaving an important gap in current research. 

In this work, we introduce SmokeBench, a comprehensive benchmark designed to assess how well multimodal large language models (MLLMs) can detect and identify wildfire smoke in images. The benchmark is structured around four progressively challenging tasks: (1) smoke classification, which evaluates whether a model can correctly recognize the presence of smoke; (2) tile-based smoke localization, which tests detection at a finer scale through cropped sub-regions; (3) grid-based smoke localization, which measures the ability to identify broader smoke regions within an image; and (4) smoke detection, which requires precise and detailed localization critical for situational awareness. Together, these tasks represent increasing levels of difficulty, covering both recognition and localization capabilities that are critical for early warning and operational wildfire management systems.

We focus our evaluation on open-source models such as Idefics2~\cite{laurenccon2024matters}, QWen2.5VL~\cite{bai2025qwen2}, InternVL3~\cite{zhu2025internvl3}, Unified-IO 2~\cite{lu2024unified}, and Grounding DINO~\cite{liu2024grounding}, while also benchmarking against closed-source counterparts including GPT-4o~\cite{hurst2024gpt} and Gemini-2.5-Pro~\cite{comanici2025gemini}. Our experiments show that while some models can classify smoke when its volume is relatively large, none demonstrate the ability to accurately localize smoke. Further, we examine the effects of smoke volume and contrast on detection performance, finding that larger volumes substantially improve predictions, whereas contrast has a comparatively limited impact. Overall, our findings reveal that current MLLMs lack the ability to detect smoke in its early stages and remain inadequate for wildfire monitoring. These limitations underline the urgent need for new techniques that enhance MLLMs’ capacity for early-stage smoke identification in wildfire scenarios.

\section{Related Work}

\noindent \textbf{Wildfire Smoke Datasets.}
The availability of datasets is crucial for advancing wildfire smoke detection.  FlgLib~\cite{dewangan2022figlib} provides around 25K  images from fixed wildfire cameras. To diversify appearance variations, Cheng et al.~\cite{cheng2019smoke} released a large set of synthetically generated smoke images for single-frame detectors.  Yan et al.~\cite{yan2022transmission} contributed pixel-level annotations (1.4K real + 4K synthetic), enabling semantic segmentation and boundary-level evaluation of amorphous, low-contrast plumes.  Other efforts focus on temporal signals: Hsu et al.~\cite{hsu2020rise} curated 12,567 video clips from industrial facilities across 19 camera views with citizen-science annotations. Lostanlen et al.~\cite{lostanlen2024scrapping} further compiled both images and videos from public wildfire-camera networks and in-house deployments, targeting early plume detection.  Operational datasets are also emerging: Pyro-SDIS~\cite{pyro-sdis} includes 33.6K YOLO-annotated images used by French fire services, and de Souza et al.~\cite{de2022automatic} released over 21K bounding-box annotated smoke images.  Together, these datasets support progress from coarse classification to fine-grained localization, though most remain domain-specific and limited in scale compared to mainstream computer vision benchmarks.

\noindent \textbf{Wildfire Smoke Detection.}
Deep learning has greatly advanced detection performance \cite{ronneberger2015unet,redmon2016you,he2017mask,li2018deep,li2023rein,chen2021channel,liu2022generalised,zheng2022towards}, with CNN- and Transformer-based detectors demonstrating strong results.  YOLO variants are widely adopted~\cite{park2024wildfire, he2025research}, sometimes combined with Mask R-CNN~\cite{mahmud2024deep} or enhanced with transformer-based U-Nets \cite{ronneberger2015unet} for multispectral segmentation~\cite{liu2024transformer}. Spatiotemporal architecture, SmokeyNet~\cite{dewangan2022figlib}, further highlight the benefits of temporal modeling. These methods achieve high accuracy under supervised settings but rely heavily on large, domain-specific annotations, limiting adaptability to diverse wildfire scenarios.

\noindent \textbf{Multimodal Large Language Models (MLLMs).} Recent advances in MLLMs \cite{liu2023visual,bai2025qwen2,laurenccon2024matters,liu2024grounding,lu2024unified,zhu2025internvl3,wu2024visionllmv2,jiao2024lumen,chen2024lion} have unified vision–language representation learning with instruction-following generation, enabling systems to ground text in pixels and reason across modalities. MLLMs including open models such as Qwen2.5-VL~\cite{bai2025qwen2}, Idefics2~\cite{laurenccon2024matters}, Grounding DINO~\cite{liu2024grounding}, Unified-IO 2~\cite{lu2024unified}, and InternVL3~\cite{zhu2025internvl3}, as well as closed-source systems like GPT-4o~\cite{hurst2024gpt} and Gemini-2.5-Pro~\cite{comanici2025gemini}, have exhibited strong zero-shot generalization across a wide range of vision–language tasks. By combining large-scale multimodal pretraining with advanced reasoning capabilities, these models achieve competitive performance for image captioning, visual question answering, visual grounding, and referring expression comprehension. Their success highlights the growing potential of MLLMs to serve as general-purpose perception and reasoning systems.

\noindent \textbf{MLLMs Benchmarks.}
Recent benchmarks~\cite{liu2024mmbench, yu2023mmvet, li2024seedbench, tong2024cambrian, robicheaux2025roboflow100, lu2025revisiting, kim2025koffvqa, li2025vidhalluc, fu2025video, schwenk2022okvqa,Zhao2026DermEVAL} have systematically evaluated MLLMs across a wide range of domains, from general visual reasoning and video understanding to open-ended knowledge grounding. These studies provide critical insights into the strengths and weaknesses of current models, highlighting challenges such as hallucination, spatio-temporal reasoning, and domain transferability. However, wildfire monitoring remain largely unexplored within this benchmarking landscape. Wildfire monitoring presents a uniquely demanding scenario: early-stage smoke plumes are often subtle, small in scale, and easily confounded with visually similar phenomena such as haze or clouds. The ability of MLLMs to reliably recognize and localize these smoke has not yet been systematically explored. Addressing this gap, our work introduces the first targeted evaluation of MLLMs on wildfire smoke detection tasks, benchmarking them to evaluate both their promise and limitations for wildfire monitoring.

\begin{figure}[t!]
  \centering
  \includegraphics[width=0.87\linewidth]{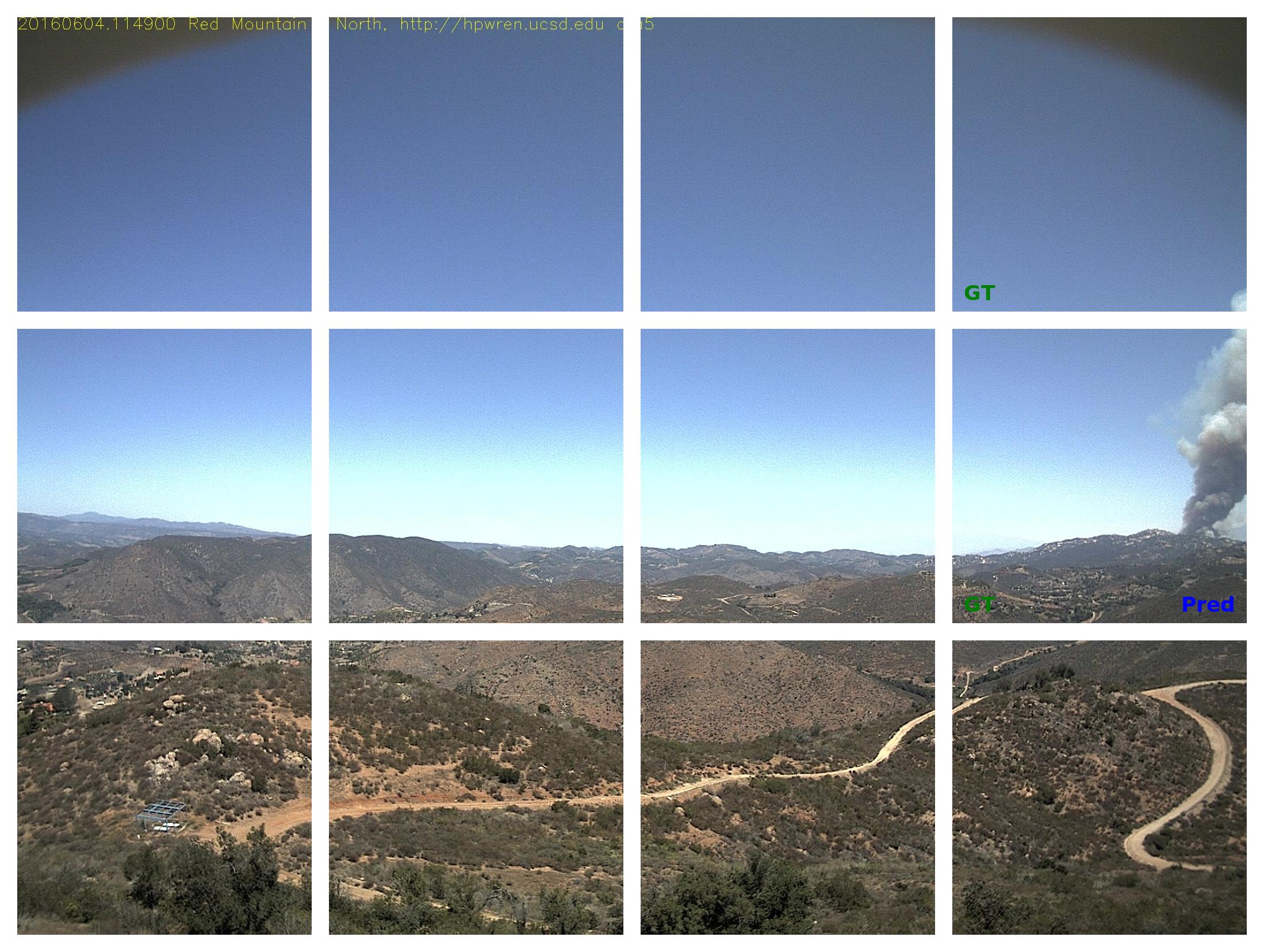}
  \caption{Example of tile-based smoke localization. The original image is cropped into a fixed $3 \times 4$ set of non-overlapping tiles. In this task, the MLLM receives each of the 12 cropped tiles as separate inputs and must independently decide whether smoke is present, thereby isolating local perception from global context. 
  }
  \label{fig:tile_result}
\end{figure}

\begin{figure}[t!]
  \centering
  \includegraphics[width=0.87\linewidth]{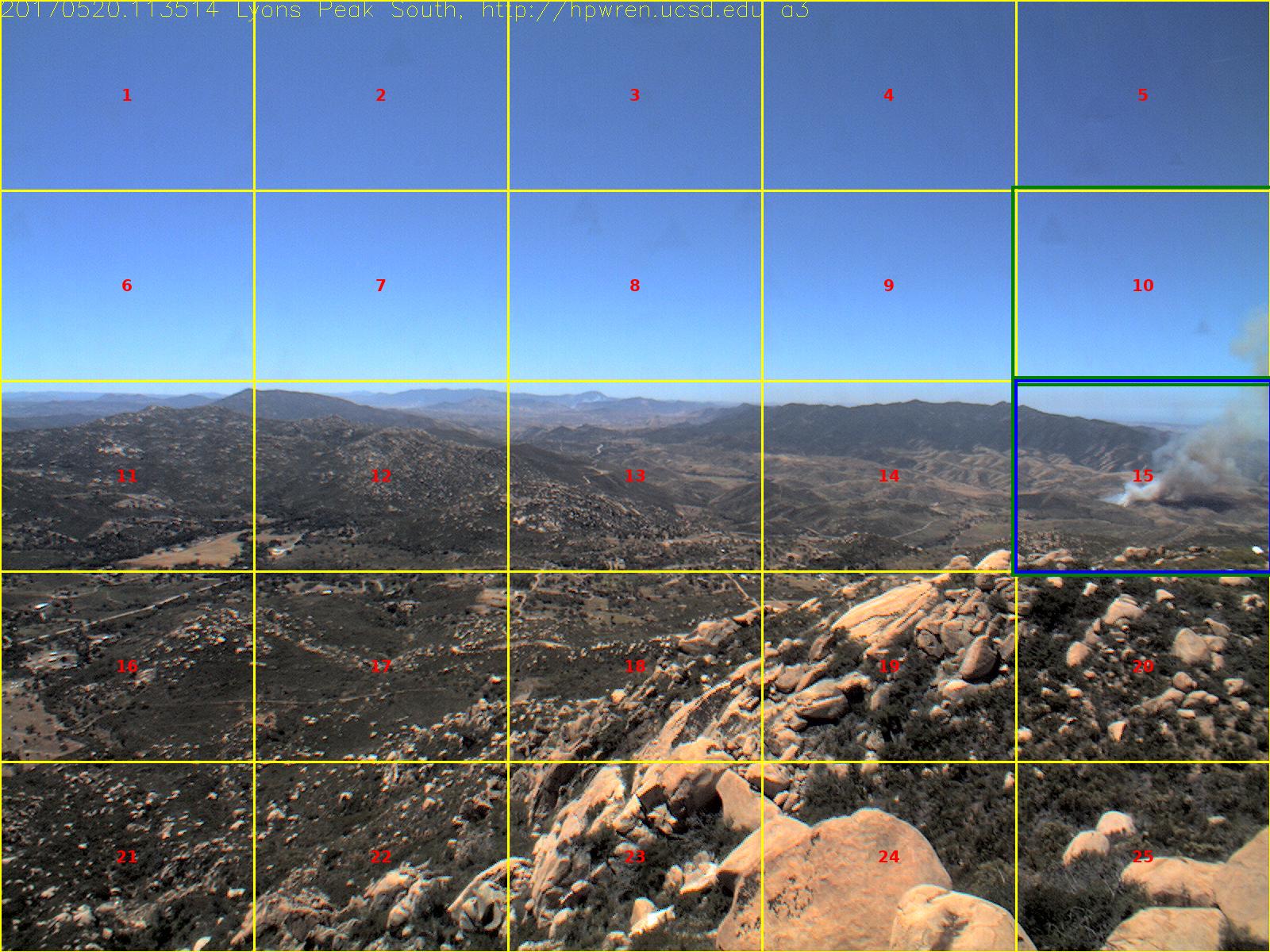}
  \caption{Example of grid-based localization on a smoke image. The image is partitioned into a $5 \times 5$ grid with unique IDs. In this task, the MLLM receives the entire image with grid overlays and is asked to identify which regions contain smoke, enabling coarse spatial reasoning without precise bounding boxes.}
  \label{fig:grid_result}
\end{figure}

\section{SmokeBench}
\label{sec:smokebench}

\subsection{Overview}

We introduce SmokeBench, a benchmark for evaluating the ability of multimodal large language models (MLLMs) to detect wildfire smoke in images. Our benchmark is built on the FIgLib dataset~\cite{dewangan2022figlib}, a large-scale image collection for wildfire smoke detection, which consists of images captured by fixed-view cameras in Southern California. We use a subset of 5,046 images with ground-truth bounding box annotations provided in \cite{dewangan2022figlib} as positive samples, along with 1,000 non-smoke images as negative samples.

Our benchmark adopts a progressive evaluation protocol to systematically evaluate MLLMs' abilities. The evaluation begins with binary smoke classification to establish coarse discrimination between smoke and non-smoke images, followed by tile-based localization, which requires identifying smoke within predefined image tiles. It then progresses to grid-based localization, a more challenging task designed to assess spatial reasoning by localizing smoke across finer partitions of the image. Finally, the evaluation culminates in smoke detection, which demands precise identification and grounding of smoke regions. 

\begin{figure*}[t]
  \centering
  \includegraphics[width=0.85\linewidth]{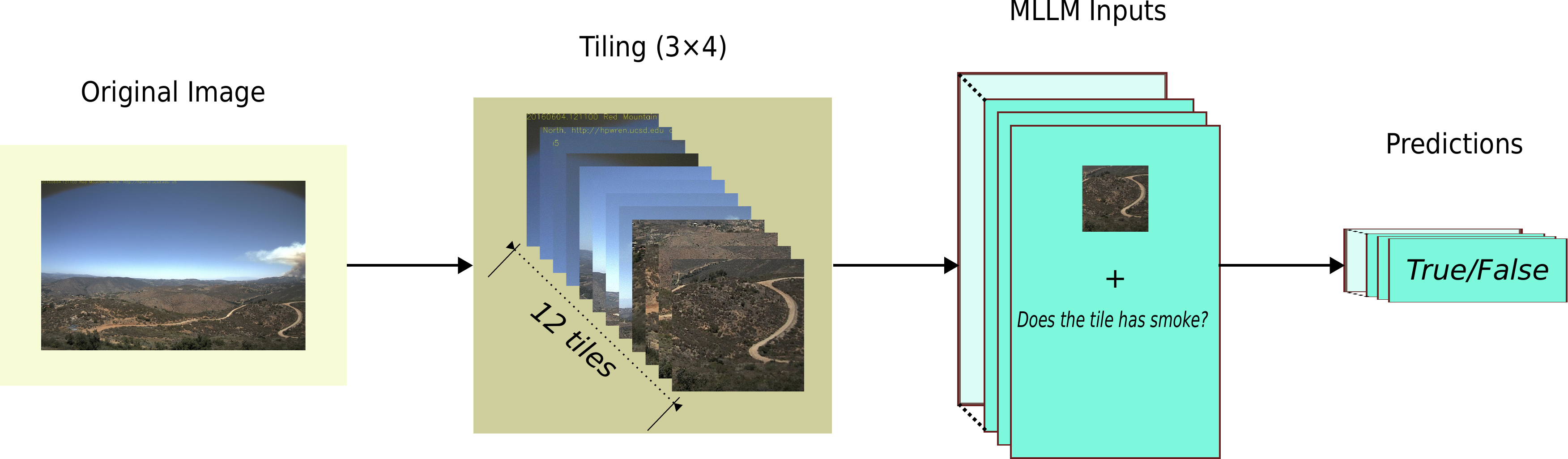}
  \caption{Pipeline of \textbf{tile classification}. The image is cropped into a fixed $3\times4$ set of tiles; the classification prompt is applied to each tile independently to determine smoke presence.}
  \label{fig:tile_pipeline}
\end{figure*}

\begin{figure*}[t]
  \centering
    \includegraphics[width=0.85\linewidth]{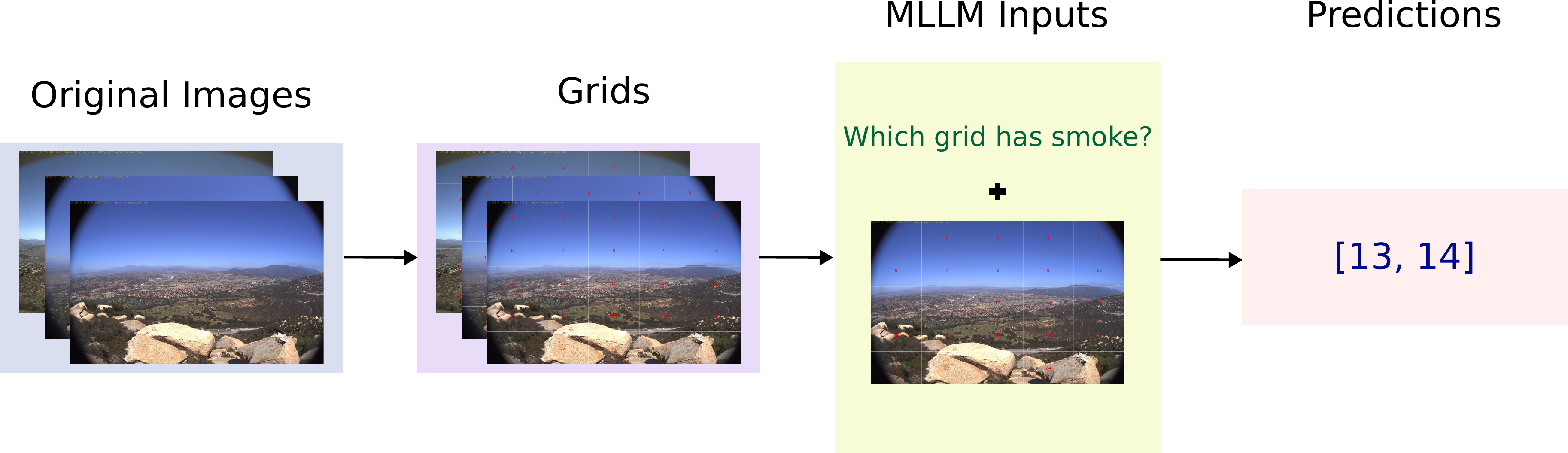}
  \caption{Pipeline of \textbf{grid-based localization}. An image is divided into a fixed $5\times5$ grid with cell IDs; the model is asked to select the cell(s) containing smoke. Ground truth is obtained by mapping bounding boxes onto grid cells.}
  \label{fig:grid_pipeline}
\end{figure*}

\subsection{Task Definitions}
\noindent \textbf{Smoke Classification.}
The first task in SmokeBench is smoke classification, where the model is asked to determine whether a given image contains wildfire smoke. Each input consists of an image paired with a natural-language instruction, and the model must output a binary decision (smoke vs. non-smoke). This task evaluates whether MLLMs can perform coarse detection of smoke. Although conceptually simple, smoke classification establishes an essential baseline: it tests the model’s ability to distinguish smoke at the image level and lays the foundation for more demanding tasks that require spatial reasoning and precise localization.

\noindent\textbf{Tile-based Smoke Localization.}
We introduce a tile-based smoke localization task. Each image is cropped into the same $m \times n$ grid of tiles, and the binary smoke classification prompt is applied independently to each tile. This setup evaluates whether a model can correctly identify smoke within localized regions when deprived of surrounding context, thereby isolating fine-grained recognition ability. Ground-truth labels are assigned to tiles that overlap annotated bounding boxes, while all others are marked negative. An example is presented in Fig.~\ref{fig:tile_result}, where the ground-truth tiles are denoted as “GT” and the model’s predictions as “Pred.” The overall pipeline is illustrated in Fig.~\ref{fig:tile_pipeline}.

\noindent\textbf{Grid-based Smoke Localization.}
We introduce a coarser but spatially aware localization task: grid-based localization. In this task, each image is partitioned into a fixed $m \times n$ grid of rectangular subregions, each assigned a unique identifier. The model is asked to predict which grid cell(s) contain smoke, thereby transforming the problem into a structured classification task over discrete spatial regions. Ground-truth labels are obtained by mapping annotated bounding boxes to grid IDs, with any cell overlapping a smoke annotation marked as positive. This formulation reduces the difficulty of precise localization while still probing the model’s ability to reason about spatial extent. An example is shown in Fig.~\ref{fig:grid_result}, where ground-truth cells are highlighted in green and model predictions are highlighted in blue. The overall pipeline is depicted in Fig.~\ref{fig:grid_pipeline}.

\noindent\textbf{Smoke Detection.}
The smoke detection task evaluates whether multimodal large language models can not only recognize the presence of smoke but also localize it within an image. To this end, the model is instructed to output bounding-box coordinates that identify the spatial extent of visible smoke plumes. The instruction is explicitly adapted to request location predictions, ensuring that the task probes grounding rather than coarse classification. Unlike dedicated object detectors, some general-purpose MLLMs (e.g., Qwen2.5-VL~\cite{bai2025qwen2}) are typically not trained to generate structured bounding-box outputs. Consequently this task is considerably more challenging: models must integrate visual recognition with spatial reasoning and structured response formatting. Performance on smoke detection therefore provides a stringent test of how well MLLMs can transfer their vision–language capabilities to grounding tasks in safety-critical domains.

\noindent Overall, these four tasks form a progressive evaluation that probes MLLMs across multiple levels of difficulty: (1) binary recognition, (2) localized classification, (3) coarse regional reasoning, and (4) structured localization. This design helps disentangle strengths and failure modes relevant to wildfire monitoring.

\section{Experiments}
In this section, we present the experimental design for evaluating MLLMs on wildfire smoke detection. We first outline the models under consideration, followed by the experimental setup. We then describe the prompting strategies for each task and conclude with the evaluation metrics used to assess model performance.

\subsection{Models}
We evaluate a diverse set of open- and closed-source multimodal large language models (MLLMs). General-purpose MLLMs include Qwen-2.5-VL-7B-Instruct and Qwen-2.5-VL-32B-Instruct~\cite{bai2025qwen2}, as well as Idefics2~\cite{laurenccon2024matters}.  These models are trained on large-scale vision--language data with instruction tuning, but lack explicit grounding supervision.  Grounding-oriented MLLMs include Unified-IO 2~\cite{lu2024unified} and InternVL3~\cite{zhu2025internvl3}, which are trained to handle visual grounding or object detection tasks, and are expected to perform better at spatial localization. We include Grounding DINO~\cite{liu2024grounding}, a specialized vision–language detector, as a strong baseline for evaluating the detection ability of MLLMs. Due to cost and resource constraints, we primarily evaluate closed-source GPT-4o~\cite{hurst2024gpt} and Gemini-2.5 Pro~\cite{comanici2025gemini} on the smoke classification task.

\subsection{Setup}

We use the bounding-box annotations provided by FIgLib~\cite{dewangan2022figlib} as ground truth. To ensure fair comparison, we design a standardized instruction prompt for each task and apply it across all models. During evaluation, models are restricted to the provided image and instruction, without access to any auxiliary context or external knowledge. All inference is conducted with the decoding temperature fixed at 0.5 to balance determinism and response diversity.

To evaluate model sensitivity, we consider two key factors: \emph{smoke size} and \emph{contrast}.  Smoke size is quantified as the pixel area of ground-truth bounding boxes.  Images are sorted by smoke area and evenly divided into five quantile-based groups, yielding ranges of  $(42,\,4356]$, $(4356,\,11232]$, $(11232,\,28565]$, $(28565,\,83157]$, and $(83157,\,2232020]$ pixels.  These correspond to the categories \textit{Very Small}, \textit{Small}, \textit{Medium}, \textit{Large}, and \textit{Very Large}, respectively.  For contrast, we adopt the Weber contrast measure \cite{peli1990contrast}, defined as
$ C = \frac{I_{\text{smoke}} - I_{\text{background}}}{I_{\text{background}}},$
where $I_{\text{smoke}}$ is the mean pixel intensity inside the annotated smoke bounding box,  and $I_{\text{background}}$ is the mean intensity of the surrounding background region.  Based on their contrast values, images are again divided into five quantile-based groups,  with ranges $(0,\,0.0246]$, $(0.0246,\,0.0549]$, $(0.0549,\,0.0878]$, $(0.0878,\,0.13]$, and $(0.13,\,0.728]$.  These correspond to \textit{Very Low}, \textit{Low}, \textit{Medium}, \textit{High}, and \textit{Very High}, respectively.  These factors provide a systematic framework for analyzing how model performance varies across both smoke scale and visibility conditions.

\subsection{Prompting Design}

\noindent \textbf{Smoke Classification.} 
This prompt is explicitly designed to enforce binary outputs, enabling straightforward and automatic evaluation.
\begin{tcolorbox} [colback=red!5!white]
  \textit{Please look at this image. Detect if the image contains smoke. 
  If you can find any smoke, return \textbf{True}. Otherwise, return \textbf{False}. 
  Do not return any other words.}
\end{tcolorbox}

\vspace{1mm}
\noindent \textbf{Tile-based Smoke Localization.} 
In this task, each image is divided into a fixed $3 \times 4$ grid of non-overlapping tiles, and the binary classification prompt is applied independently to each region. By evaluating tiles in isolation, the task probes whether models can recognize smoke based solely on local visual cues, without leveraging global context or scene-level correlations.

\vspace{1mm}
\noindent \textbf{Grid-based Smoke Localization.} 
For region-level localization, the prompt specifies a JSON output format for consistency:
\begin{tcolorbox} [colback=red!5!white]
\textit{Please look at this image, which is divided into several numbered regions 
(from left to right, top to bottom). Please output the numbered regions that contain smoke 
in JSON format as a list of dicts like [\{"region": 1\}, \{"region": 2\}]. 
If you cannot find any smoke, return an empty list []. without other words}
\end{tcolorbox}

\vspace{1mm}
\noindent \textbf{Smoke Detection.} 
For smoke detection, models are required to produce explicit bounding-box coordinates corresponding to the location of wildfire smoke.
\begin{tcolorbox} [colback=red!5!white]
 \textit{Detect all smoke and output bounding boxes in the format  [[$x_1, y_1, x_2, y_2$], [$x_1, y_1, x_2, y_2$]].  If you cannot find any smoke, return an empty list [] without other words. Do not return dictionaries.}
\end{tcolorbox}

\noindent In practice, we observed that some MLLMs occasionally returned inconsistent JSON structures, such as dictionaries instead of lists. To address this issue, the prompt explicitly constrains the required output format.

\subsection{Metrics}
We report results using two standard metrics:

\noindent \textbf{Accuracy}: The proportion of correctly classified images over the total images with ground truth. The MLLM is prompted to output a categorical answer \texttt{True} or \texttt{False} per image; we map \texttt{True}$\to 1$, \texttt{False}$\to 0$, and compute: $ \text{Accuracy}=\frac{1}{N}\sum_{i=1}^{N}\mathbf{1}\big[\hat{y}_i=y_i\big], $ where $y_i\in\{0,1\}$ is the ground-truth presence label (smoke/non-smoke) and $\hat{y}_i\in\{0,1\}$ is the model’s direct categorical output.

\vspace{1mm}
\noindent \textbf{Mean Intersection over Union (mIoU)}: 
The primary metric for evaluating detection quality is the mean Intersection over Union (mIoU). For a single image, the Intersection over Union (IoU) is defined as 
 $ \text{IoU} = \frac{|P \cap G|}{|P \cup G|}, $ where $P$ and $G$ denote the predicted and ground-truth smoke regions. The mean IoU is then computed by averaging IoU scores across all test images, providing a comprehensive measure of overall localization accuracy.  

\begin{table*}[t]
\centering
\small
\begin{tabular*}{0.85\textwidth}{@{\extracolsep{\fill}}lccccccc}
\toprule
\multirow{2}{*}{Model} & \multirow{2}{*}{Size} &
\multicolumn{6}{c}{Accuracy on Different Area Bins} \\
\cmidrule(lr){3-8}
& & Very Small & Small & Medium & Large & Very Large & Overall \\
\midrule
\multicolumn{8}{c}{Open-source models} \\
\midrule
Idefics2        & 8B  & \textbf{0.482} & \textbf{0.596} & \textbf{0.598} & \textbf{0.575} & \textbf{0.709} & \textbf{0.592} \\
Qwen2.5-VL      & 7B  & 0.098 & 0.232 & 0.423 & 0.463 & \textbf{0.686} & 0.380 \\
Qwen2.5-VL      & 32B & 0.041 & 0.103 & 0.222 & 0.284 & 0.569 & 0.244 \\
InternVL3       & 14B & 0.090 & 0.262 & 0.376 & 0.432 & 0.636 & 0.359 \\
\midrule
\multicolumn{8}{c}{\textit{Closed-source models}} \\
\midrule
Gemini-2.5 Pro  & –   & 0.441 & 0.608 & 0.713 & 0.773 & 0.882 & 0.683 \\
GPT-4o          & –   & \textbf{0.529} & \textbf{0.727} & \textbf{0.770} & \textbf{0.831} & \textbf{0.931} & \textbf{0.758} \\
\bottomrule
\end{tabular*}
\caption{Classification accuracy across quantile-based bins of ground-truth smoke area 
(Very Small–Very Large). Bold values indicate the strongest model in each group 
(open-source vs. closed-source). Results show that detection accuracy improves with larger smoke regions, 
and while Idefics2 leads among open-source models, GPT-4o is consistently best overall.}
\label{tab:area-acc-classification}
\end{table*}

\begin{table*}[t]
\centering
\small
\begin{tabular*}{0.85\textwidth}{@{\extracolsep{\fill}}lccccccc}
\toprule
\multirow{2}{*}{Model} & \multirow{2}{*}{Size} &
\multicolumn{6}{c}{Accuracy on Different Contrast Groups} \\
\cmidrule(lr){3-8}
& & Very Low & Low & Medium & High & Very High & Overall \\
\midrule
\multicolumn{8}{c}{\textit{Open-source models}} \\
\midrule
Idefics2        & 8B  & \textbf{0.607} & \textbf{0.589} & \textbf{0.616} & \textbf{0.558} & \textbf{0.589} & \textbf{0.592} \\
Qwen2.5-VL      & 7B  & 0.289 & 0.327 & 0.395 & 0.417 & 0.473 & 0.380 \\
Qwen2.5-VL      & 32B & 0.204 & 0.224 & 0.251 & 0.249 & 0.291 & 0.244 \\
InternVL3       & 14B & 0.269 & 0.304 & 0.368 & 0.401 & 0.453 & 0.359 \\
\midrule
\multicolumn{8}{c}{\textit{Closed-source models}} \\
\midrule
Gemini-2.5 Pro  & –   & 0.588 & 0.629 & 0.717 & 0.696 & 0.786 & 0.683 \\
GPT-4o          & –   & \textbf{0.739} & \textbf{0.703} & \textbf{0.774} & \textbf{0.792} & \textbf{0.781} & \textbf{0.758} \\
\bottomrule
\end{tabular*}
\caption{Classification accuracy across quantile-based contrast groups 
(Very Low–Very High). Bold values indicate the strongest model in each group 
(open-source vs. closed-source). Results show that contrast has a weaker effect on accuracy 
compared to smoke area size.}
\label{tab:contrast-acc-classification}
\end{table*}

\section{Results}
\label{sec:results}
In this section, we present a comprehensive evaluation of multimodal large language models (MLLMs) on wildfire smoke detection. We analyze performance with respect to smoke area and contrast, providing a systematic comparison across settings.

\subsection{Smoke Classification}

We evaluate the smoke classification ability of both open-source and closed-source MLLMs, measured by accuracy. Table~\ref{tab:area-acc-classification} reports classification accuracy across smoke area bins for both model categories. The results show a strong dependency on the spatial extent of smoke regions. For the \textit{very small} and \textit{small} categories, only Idefics2 achieves non-trivial accuracy, whereas Qwen2.5-VL and InternVL3 struggle to detect smoke and often fail. As smoke size increases, performance improves steadily: Qwen2.5-VL and InternVL3 surpass 0.6 accuracy in the \textit{very large} bin, and Idefics2 maintains consistently higher performance relative to its open-source peers. Among the closed-source models, Gemini-2.5 Pro performs well across all bins, but GPT-4o achieves the strongest results, reaching over 0.9 accuracy for \textit{very large} smoke regions and outperforming all competitors overall. These results show that while MLLMs can detect large, visually prominent smoke plumes, they remain unreliable for small or early-stage smoke.

Table~\ref{tab:contrast-acc-classification} examines the impact of Weber contrast between smoke and its background regions. Idefics2 is stable across contrast bins, while GPT-4o again dominates, achieving the highest accuracy in every group. We find that while higher contrast generally correlates with better performance, the magnitude of improvement is relatively modest (often only $0.03$--$0.05$ from low to high contrast for most models). Although contrast has some effect, it is less influential than the spatial scale of smoke. 

\begin{table}[t]
\centering
\small
\begin{tabular*}{0.6\linewidth}{@{\extracolsep{\fill}}lc}
\toprule
Model & Accuracy \\
\midrule
Idefics2 (8B)         & 0.498 \\
Qwen2.5-VL (7B)   & 0.982 \\
Qwen2.5-VL (32B)  & 0.983 \\
InternVL3 (14B)   & \textbf{0.985} \\
\bottomrule
\end{tabular*}
\caption{Classification accuracy on negative samples (images without smoke). 
Higher is better; this reflects performance on smoke-free images (i.e., specificity/true negative rate).}
\label{tab:negative-acc}
\end{table}

We present performance of MLLMs on negative samples, i.e., images without smoke, in Table~\ref{tab:negative-acc}. Qwen2.5-VL (7B and 32B) and InternVL3 achieve high accuracy (above 0.98), demonstrating strong specificity and reliable identification of smoke-free images. In contrast, Idefics2 performs near random guessing (0.498). This phenomenon explains why Idefics2 appears to perform better on positive smoke images: its random predictions occasionally coincide with the ground truth, resulting in inflated accuracy that superficially exceeds other open-source models. In summary, these classification results reveal that current MLLMs are unable to provide robust smoke classification in the conditions most relevant for early wildfire warning, limiting their real-world applicability.

\subsection{Tile-based Smoke Localization}

\begin{table*}[t!]
\centering
\small
\begin{tabular*}{0.8\textwidth}{@{\extracolsep{\fill}}lccccccc}
\toprule
\multirow{2}{*}{Model} & \multirow{2}{*}{Size} &
\multicolumn{6}{c}{mIoU on Different Area Bins} \\
\cmidrule(lr){3-8}
& & Very Small & Small & Medium & Large & Very Large & Overall \\
\midrule
Idefics2          & 8B  & \textbf{0.348} & \textbf{0.505} & 0.487 & \textbf{0.530} & 0.507 & \textbf{0.475} \\
Qwen2.5-VL (7B)   & 7B  & 0.311 & 0.479 & \textbf{0.485} & 0.485 & \textbf{0.520} & 0.456 \\
Qwen2.5-VL (32B)  & 32B & \textbf{0.348} & 0.466 & 0.422 & 0.456 & 0.483 & 0.435 \\
\bottomrule
\end{tabular*}
\caption{Tile-based smoke localization performance (area-binned). 
Each image is cropped into $3\times 4$ tiles (12 regions). 
Results report mean IoU averaged within five quantile-based smoke area bins.}
\label{tab:area-iou-tiles}
\end{table*}

\begin{table*}[t!]
\centering
\small
\begin{tabular*}{0.8\textwidth}{@{\extracolsep{\fill}}lccccccc}
\toprule
\multirow{2}{*}{Model} & \multirow{2}{*}{Size} &
\multicolumn{6}{c}{mIoU on Different Contrast Groups} \\
\cmidrule(lr){3-8}
& & Very Low & Low & Medium & High & Very High & Overall \\
\midrule
Idefics2          & 8B  & \textbf{0.416} & 0.440 & \textbf{0.480} & \textbf{0.477} & \textbf{0.563} & \textbf{0.475} \\
Qwen2.5-VL (7B)   & 7B  & 0.376 & \textbf{0.458} & \textbf{0.482} & 0.478 & 0.485 & 0.456 \\
Qwen2.5-VL (32B)  & 32B & 0.380 & 0.420 & 0.450 & 0.440 & 0.485 & 0.435 \\
\bottomrule
\end{tabular*}
\caption{Tile-based smoke localization performance (contrast-binned). 
Using the same $3\times 4$ tiling setup, results are grouped by Weber contrast (Very Low–Very High).}
\label{tab:contrast-iou-tiles}
\end{table*}

We evaluate the tile-based smoke classification, which simplifies the problem further by cropping each image into a $3 \times 4$ grid of tiles and asking models to classify each tile independently. As shown in Table~\ref{tab:area-iou-tiles} and Table~\ref{tab:contrast-iou-tiles}, both Idefics2 and QWen2.5VL achieve relatively strong performance on this task compared to the grid and bounding-box settings. This demonstrates that MLLMs are better suited to coarse-grained classification than precise spatial localization. Notably, Idefics2 and QWen2.5VL-32B show comparable results, while QWen2.5VL-7B consistently underperforms, highlighting the effect of model scale.

Nevertheless, the improvement from tile-based smoke localization is conditional. As illustrated in Table~\ref{tab:area-iou-tiles} , both Idefics2 and QWen2.5VL models only begin to perform reliably when smoke areas reach the \textit{Medium} size bin. At smaller scales, the models fail regardless of Weber contrast, showing that area size remains the dominant factor. While contrast does influence results—higher contrast generally correlates with better mIoU—the gains are modest and saturate quickly once smoke regions are large enough. Interestingly, Idefics2 appears to maintain relatively stable performance across different contrast groups, whereas QWen2.5VL-32B exhibits more fluctuation, which may reflect architectural differences in robustness.

In summary, tile-based smoke localization represents the most promising of the four tasks evaluated, as it reduces contextual distractions and simplifies the decision-making process. However, even this approach fails to provide reliable detection when smoke is small or early-stage. This limitation further underscores the gap between current MLLMs and the requirements of real-world wildfire monitoring systems, where early and precise detection is critical.

\subsection{Grid-based Smoke Localization}

\begin{table*}[t]
\centering
\small
\begin{tabular*}{0.8\textwidth}{@{\extracolsep{\fill}}lccccccc}
\toprule
\multirow{2}{*}{Model} & \multirow{2}{*}{Size} &
\multicolumn{6}{c}{mIoU on Different Area Bins} \\
\cmidrule(lr){3-8}
& & Very Small & Small & Medium & Large & Very Large & Overall \\
\midrule
Idefics2     & 8B  & 0.029 & 0.025 & 0.033 & 0.054 & 0.067 & 0.042 \\
Qwen2.5-VL   & 7B  & 0.000 & 0.000 & 0.000 & 0.000 & 0.000 & 0.000 \\
Qwen2.5-VL   & 32B & \textbf{0.057} & \textbf{0.155} & \textbf{0.242} & \textbf{0.226} & \textbf{0.239} & \textbf{0.184} \\
InternVL3   & 14B & 0.001 & 0.017 & 0.057 & 0.095 & 0.114 & 0.057 \\
\bottomrule
\end{tabular*}
\caption{Grid-based localization performance (area-binned). 
Images are divided into $5\times 5$ grids (25 cells), and models predict which cells contain smoke. 
Results are averaged across five area bins of ground-truth smoke regions.}
\label{tab:area-iou-grid}
\end{table*}

\begin{table*}[t]
\centering
\small
\begin{tabular*}{0.8\textwidth}{@{\extracolsep{\fill}}lccccccc}
\toprule
\multirow{2}{*}{Model} & \multirow{2}{*}{Size} &
\multicolumn{6}{c}{mIoU on Different Contrast Groups} \\
\cmidrule(lr){3-8}
& & Very Low & Low & Medium & High & Very High & Overall \\
\midrule
Idefics2          & 8B  & 0.036 & 0.042 & 0.040 & 0.048 & 0.043 & 0.042 \\
Qwen2.5-VL (7B)   & 7B  & 0.000 & 0.000 & 0.000 & 0.000 & 0.000 & 0.000 \\
Qwen2.5-VL (32B)  & 32B & \textbf{0.155} & \textbf{0.157} & \textbf{0.189} & \textbf{0.202} & \textbf{0.215} & \textbf{0.184} \\
InternVL3 (14B)   & 14B & 0.042 & 0.041 & 0.055 & 0.062 & 0.084 & 0.057 \\
\bottomrule
\end{tabular*}
\caption{Grid-based localization performance (contrast-binned). 
Images are divided into $5\times 5$ grids (25 cells), and results are grouped by Weber contrast levels (Very Low–Very High).}
\label{tab:contrast-iou-grid}
\end{table*}

To reduce the difficulty of structured localization, we designed the grid-based localization task, which requires models to identify sub-regions of an image rather than producing exact bounding-box coordinates. Compared with the bounding box setting, the grid task yields better but still limited results. As shown in Table~\ref{tab:area-iou-grid} and Table~\ref{tab:contrast-iou-grid}, performance again depends heavily on smoke size. In particular, QWen2.5VL-32B starts to achieve non-trivial mIoU values only when smoke areas reach the \textit{Medium} size bin. Smaller models, such as QWen2.5VL-7B, or other general-purpose MLLMs like Idefics2 and InternVL3, remain close to random guessing across most bins. These results suggest that current MLLMs only become useful when smoke is already visually prominent in the image, which corresponds to relatively late stages of wildfire development.

The effect of Weber contrast is comparatively minor. While higher contrast does provide a slight boost in performance (Table~\ref{tab:contrast-iou-grid}), the improvement is not sufficient to compensate for the difficulties posed by small smoke regions. These findings indicate that while the grid task does make the problem more tractable than bounding-box localization, current MLLMs still lack the robustness and fine-grained spatial reasoning required for reliable real-world wildfire detection. In practice, their detection capability only becomes apparent when wildfires have already grown into potentially uncontrollable scenarios.

\subsection{Smoke Detection}

\begin{table}[t!]
\centering
\small
\begin{tabular*}{0.65\linewidth}{@{\extracolsep{\fill}}lc}
\toprule
Model & mIoU \\
\midrule
Idefics2          & 0.000 \\
Qwen2.5-VL (7B)   & 0.000 \\
Qwen2.5-VL (32B)  & 0.000 \\
InternVL3 (14B)   & 0.000 \\
Unified-IO 2              & 0.092 \\
GroundingDINO     & \textbf{0.245} \\
\midrule
\textcolor{gray}{YOLOv8n} & \textcolor{gray}{0.773} \\
\bottomrule
\end{tabular*}
\vspace{-1mm}
\caption{Mean IoU scores for smoke detection. Values are low overall, with GroundingDINO achieving the highest performance.}
\label{tab:bbox}
\end{table}

We evaluate whether MLLMs are capable of structured localization through bounding-box prediction, measured by mean IoU. The results in Table~\ref{tab:bbox} show that all general-purpose MLLMs, including Idefics2, QWen2.5VL-7B, QWen2.5VL-32B, and InternVL3-14B, fail completely on this task, with IoU values close to zero. Among grounding models, Unified-IO 2—explicitly trained for grounding and object localization—achieves only a marginal score of 0.092. Grounding DINO, a state-of-the-art visual grounding detector, performs somewhat better at 0.245, which is still lower than specialized smoke detection models such as YOLOv8n \cite{chen2025recognition}.  These results highlight two important observations. First, general-purpose MLLMs, despite their strong performance on a wide variety of vision–language benchmarks, show virtually no capability for smoke localization when precise bounding boxes are required. Second, even models trained explicitly for grounding or object detection struggle in this domain. The likely reasons lie in the inherent properties of wildfire smoke: it often has amorphous and transparent boundaries, irregular shapes, and low contrast with its background. Together, these factors make smoke much harder to localize than conventional rigid objects such as people, animals, or vehicles. Thus, bounding-box localization of smoke remains an unsolved and extremely challenging problem.
\vspace{-1mm}

\section{Conclusion}

We introduced a benchmark, called SmokeBench, to evaluate the ability of multimodal large language models (MLLMs) to recognize and localize wildfire smoke in images.  We conducted a systematic evaluation of MLLMs on wildfire smoke detection across four progressively challenging tasks: binary smoke classification, tile-based smoke localization, grid-based  smoke localization, and bounding-box smoke detection. Our analysis revealed three consistent patterns. First, smoke area size is the dominant factor: all models fail to detect small or early-stage smoke, though performance improves when smoke becomes visually prominent. Second, contrast between smoke and background offers only marginal gains, indicating that scale rather than visibility is the primary bottleneck. Third, while coarse-grained tasks such as tile-based smoke localization expose some potential, current MLLMs are essentially incapable of precise localization. These results demonstrate that wildfire smoke detection is qualitatively different from conventional object detection, requiring sensitivity to faint, amorphous, and low-contrast structures that current models are ill-equipped to capture. Consequently, existing MLLMs remain inadequate for providing reliable early-warning signals in practical wildfire surveillance systems. Future research should explore domain-adapted training strategies, finer-grained supervision, and the integration of spatiotemporal signals to bridge this gap and move toward models capable of supporting safety-critical early detection.

\paragraph{Acknowledgement} This research was in part supported by the ANU-Optus Bushfire Research Centre of Excellence.

\clearpage
{
    \newpage
    \bibliographystyle{ieeenat_fullname}
    \bibliography{main}
}

\end{document}